\documentclass[10pt,letterpaper,twocolumn]{article}
\usepackage[latin9]{inputenc}
\usepackage{xcolor}
\usepackage{amsmath}
\usepackage{amssymb}
\usepackage{graphicx}
\usepackage[unicode=true,
 bookmarks=false,
 breaklinks=true,pdfborder={0 0 1},backref=section,colorlinks=false]
 {hyperref}
\hypersetup{
 pagebackref=true,letterpaper=true,colorlinks}

\makeatletter

\providecommand{\tabularnewline}{\\}

\usepackage{cvpr}
\usepackage{times}
\usepackage{epsfig}
\usepackage{graphicx}
\usepackage{microtype}
\usepackage{balance}
\usepackage{flushend}

\usepackage[linesnumbered,algoruled,boxed,lined]{algorithm2e}
\usepackage{siunitx}

\cvprfinalcopy 

\def\cvprPaperID{6600} 

\ifcvprfinal\fi

\makeatother

\begin{document}
\title{Hierarchical Conditional Relation Networks for Video Question Answering}
\author{Thao Minh Le, Vuong Le, Svetha Venkatesh, and Truyen Tran\\
Applied Artificial Intelligence Institute, Deakin University, Australia\\
 \texttt{\small{}\{lethao,vuong.le,svetha.venkatesh,truyen.tran\}@deakin.edu.au}}

\maketitle
\global\long\def\ModelName{\text{HCRN}}%
\global\long\def\UnitName{\text{CRN}}%
\global\long\def\Problem{\text{VideoQA}}%

\thispagestyle{empty}
\begin{abstract}
Video question answering ($\Problem$) is challenging as it requires
modeling capacity to distill dynamic visual artifacts and distant
relations and to associate them with linguistic concepts. We introduce
a general-purpose reusable neural unit called Conditional Relation
Network ($\UnitName$) that serves as a building block to construct
more sophisticated structures for representation and reasoning over
video. $\UnitName$ takes as input an array of tensorial objects and
a conditioning feature, and computes an array of encoded output objects.
Model building becomes a simple exercise of replication, rearrangement
and stacking of these reusable units for diverse modalities and contextual
information. This design thus supports high-order relational and multi-step
reasoning. The resulting architecture for $\Problem$ is a $\UnitName$
hierarchy whose branches represent sub-videos or clips, all sharing
the same question as the contextual condition. Our evaluations on
well-known datasets achieved new SoTA results, demonstrating the impact
of building a general-purpose reasoning unit on complex domains such
as $\Problem$. 

\end{abstract}

\section{Introduction}

Answering natural questions about a video is a powerful demonstration
of cognitive capability. The task involves acquisition and manipulation
of spatio-temporal visual representations guided by the compositional
semantics of the linguistic cues \cite{gao2018motion,lei2018tvqa,li2019beyond,song2018explore,tapaswi2016movieqa,wang2018movie}.
As questions are potentially unconstrained, $\Problem$ requires deep
modeling capacity to encode and represent crucial video properties
such as object permanence, motion profiles, prolonged actions, and
varying-length temporal relations in a hierarchical manner. For $\Problem$,
the visual representations should ideally be question-specific and
answer-ready. 

The current approach toward modeling videos for QA is to build neural
architectures in which each sub-system is either designed for a specific
tailor-made purpose or for a particular data modality. Because of
this specificity, such hand crafted architectures tend to be non-optimal
for changes in data modality \cite{lei2018tvqa}, varying video length
\cite{na2017read} or question types (such as frame QA \cite{li2019beyond}
versus action count \cite{fan2019heterogeneous}). This has resulted
in proliferation of heterogeneous networks.

\begin{figure*}
\begin{minipage}[c][1\totalheight][b]{0.48\textwidth}%
\begin{center}
\includegraphics[width=1\textwidth]{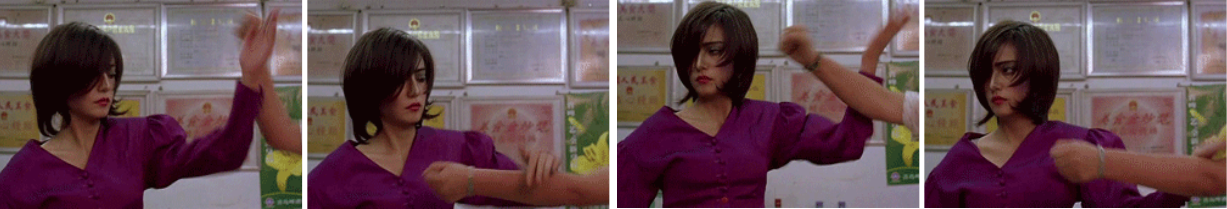}
\par\end{center}
\vspace{-3mm}
{\small{}(a) Question: What does the girl do 9 times?}{\small\par}

{\small{}\qquad{}Baseline: }\textcolor{red}{\small{}walk}{\small\par}

{\small{}\qquad{}$\ModelName$: }\textcolor{green}{\small{}blocks
a person's punch}{\small\par}

{\small{}\qquad{}Ground truth: }\textcolor{brown}{\small{}blocks
a person's punch}{\small\par}%
\end{minipage}\hfill{}%
\begin{minipage}[c][1\totalheight][b]{0.48\textwidth}%
\begin{center}
\includegraphics[width=1\textwidth]{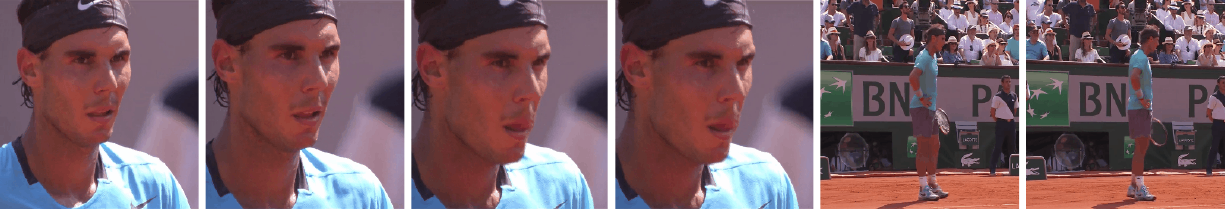}
\par\end{center}
\vspace{-3mm}
{\small{}(b) Question: What does the man do before turning body to
left?}{\small\par}

{\small{}\qquad{}Baseline: }\textcolor{red}{\small{}pick up the man's
hand}{\small\par}

{\small{}\qquad{}$\ModelName$: }\textcolor{green}{\small{}breath}{\small\par}

{\small{}\qquad{}Ground truth: }\textcolor{brown}{\small{}breath}{\small\par}%
\end{minipage}\vspace{3mm}

\caption{Example questions for which frame relations are key toward correct
answers. {\small{}(a) }\emph{\small{}Near-term frame relations }{\small{}are
required for counting of fast actions. (b) }\emph{\small{}Far-term
frame relations }{\small{}connect the actions in long transition.}
$\protect\ModelName$ with the ability to model hierarchical conditional
relations handles successfully, while baseline struggles. See more
examples in supplemental materials.\label{fig:qualitative}}
\end{figure*}

In this work we propose a general-purpose \emph{reusable neural unit}
called Conditional Relation Network ($\UnitName)$ that encapsulates
and transforms an array of objects into a new array conditioned on
a contextual feature. The unit computes sparse high-order relations
between the input objects, and then modulates the encoding through
a specified context (See Fig.~\ref{fig:Illustration-of-Multiscales}).
The flexibility of $\UnitName$ and its encapsulating design allow
it to be replicated and layered to form deep hierarchical conditional
relation networks ($\ModelName$) in a straightforward manner. The
stacked units thus provide contextualized refinement of relational
knowledge from video objects -- in a stage-wise manner it combines
appearance features with clip activity flow and linguistic context,
and follows it by integrating in context from the whole video motion
and linguistic features. The resulting $\ModelName$ is homogeneous,
agreeing with the design philosophy of networks such as InceptionNet
\cite{szegedy2015going}, ResNet \cite{he2016deep} and FiLM \cite{perez2018film}.

The hierarchy of the $\UnitName$s are as follows -- at the lowest
level, the $\UnitName$s encode the relations \emph{between} frame
appearance in a clip and integrate the \emph{clip motion as context};
this output is processed at the next stage by $\UnitName$s that now
integrate in the \emph{linguistic context}; in the following stage,
the $\UnitName$s capture the relation \emph{between} the clip encodings,
and integrate in \emph{video motion as context}; in the final stage
the $\UnitName$ integrates the video encoding with the linguistic
feature as context (See Fig.~\ref{fig:Overview-of-Network}). By
allowing the $\UnitName$s to be stacked hierarchically, the model
naturally supports modeling hierarchical structures in video and relational
reasoning; by allowing appropriate context to be introduced in stages,
the model handles multimodal fusion and multi-step reasoning. For
long videos further levels of hierarchy can be added enabling encoding
of relations between distant frames.

We demonstrate the capability of $\ModelName$ in answering questions
in major $\Problem$ datasets. The hierarchical architecture with
four-layers of $\UnitName$ units achieves favorable answer accuracy
across all $\Problem$ tasks. Notably, it performs consistently well
on questions involving either appearance, motion, state transition,
temporal relations, or action repetition demonstrating that the model
can analyze and combine information in all of these channels. Furthermore
$\ModelName$ scales well on longer length videos simply with the
addition of an extra layer. Fig.~\ref{fig:qualitative} demonstrates
several representative cases those were difficult for the baseline
of flat visual-question interaction but can be handled by our model.

Our model and results demonstrate the impact of building general-purpose
neural reasoning units that support native multimodality interaction
in improving robustness and generalization capacities of $\Problem$
models.

\section{Related Work \label{sec:Related-Work}}

Our proposed $\ModelName$ model advances the development of $\Problem$
by addressing two key challenges: (1) Efficiently representing videos
as amalgam of complementing factors including appearance, motion and
relations, and (2) Effectively allows the interaction of such visual
features with the linguistic query.

\textbf{Spatio-temporal video representation} is traditionally done
by variations of recurrent networks (RNNs) among which many were used
for $\Problem$ such as recurrent encoder-decoder \cite{zhu2017uncovering,zhao2019long},
bidirectional LSTM \cite{kim2017deepstory} and two-staged LSTM \cite{zeng2017leveraging}.
To increase the memorizing ability, external memory can be added to
these networks \cite{gao2018motion,zeng2017leveraging}. This technique
is more useful for videos that are longer \cite{xu2016msr} and with
more complex structures such as movies \cite{tapaswi2016movieqa}
and TV programs \cite{lei2018tvqa} with extra accompanying channels
such as speech or subtitles. On these cases, memory networks \cite{kim2017deepstory,na2017read,wang2019holistic}
were used to store multimodal features \cite{wang2018movie} for later
retrieval. Memory augmented RNNs can also compress video into heterogenous
sets \cite{fan2019heterogeneous} of dual appearance/motion features.
While in RNNs, appearance and motion are modeled separately, 3D and
2D/3D hybrid convolutional operators \cite{tran2018closer,qiu2017learning}
intrinsically integrates spatio-temporal visual information and are
also used for $\Problem$ \cite{jang2017tgif,li2019beyond}. Multiscale
temporal structure can be modeled by either mixing short and long
term convolutional filters \cite{wu2019long} or combining pre-extracted
frame features non-local operators \cite{tang2018non,li2017temporal}.
Within the second approach, the TRN network \cite{zhou2018temporal}
demonstrates the role of temporal frame relations as an another important
visual feature for video reasoning and VideoQA \cite{le2019learning2reason}.
Relations of predetected objects were also considered in a separate
processing stream \cite{jin2019multi} and combined with other modalities
in late-fusion \cite{singh2019spatio}. Our $\ModelName$ model emerges
on top of these trends by allowing all three channels of video information
namely appearance, motion and relations to iteratively interact and
complement each other in every step of a hierarchical multi-scale
framework.

Earlier attempts for generic multimodal fusion for visual reasoning
includes bilinear operators, either applied directly \cite{kim2018bilinear}
or through attention \cite{kim2018bilinear,yu2017multi}. While these
approaches treat the input tensors equally in a costly joint multiplicative
operation, $\ModelName$ separates conditioning factors from refined
information, hence it is more efficient and also more flexible on
adapting operators to conditioning types.

Temporal hierarchy has been explored for video analysis \cite{lienhart1999abstracting},
most recently with recurrent networks \cite{pan2016hierarchical,baraldi2017hierarchical}
and graph networks \cite{mao2018hierarchical}. However, we believe
we are the first to consider hierarchical interaction of multi-modalities
including linguistic cues for $\Problem$.

\textbf{Linguistic query--visual feature interaction} \textbf{in
VideoQA} has traditionally been formed as a visual information retrieval
task in a common representation space of independently transformed
question and referred video \cite{zeng2017leveraging}. The retrieval
is more convenient with heterogeneous memory slots \cite{fan2019heterogeneous}.
On top of information retrieval, co-attention between the two modalities
provides a more interactive combination \cite{jang2017tgif}. Developments
along this direction include attribute-based attention \cite{ye2017video},
hierarchical attention \cite{liang2018focal,zhao2018multi,zhao2017video},
multi-head attention \cite{kim2018multimodal,li2019learnable}, multi-step
progressive attention memory \cite{kim2019progressive} or combining
self-attention with co-attention \cite{li2019beyond}. For higher
order reasoning, question can interact iteratively with video features
via episodic memory or through switching mechanism \cite{yang2019question}.
Multi-step reasoning for VideoQA is also approached by \cite{xu2017video}
and \cite{song2018explore} with refined attention.

Unlike these techniques, our $\ModelName$ model supports conditioning
video features with linguistic clues as a context factor in every
stage of the multi-level refinement process. This allows linguistic
cue to involve earlier and deeper into video presentation construction
than any available methods.

\textbf{Neural building blocks} - Beyond the VideoQA domain, $\UnitName$
unit shares the idealism of uniformity in neural architecture with
other general purpose neural building blocks such as the block in
InceptionNet \cite{szegedy2015going}, Residual Block in ResNet \cite{he2016deep},
Recurrent Block in RNN, conditional linear layer in FiLM \cite{perez2018film},
and matrix-matrix-block in neural matrix net \cite{do2018learning}.
Our $\UnitName$ departs significantly from these designs by assuming
an array-to-array block that supports conditional relational reasoning
and can be reused to build networks of other purposes in vision and
language processing.

\section{Method \label{sec:Method}}

\begin{figure}
\begin{centering}
\vspace{-4mm}
\includegraphics[width=0.88\columnwidth,height=0.92\columnwidth]{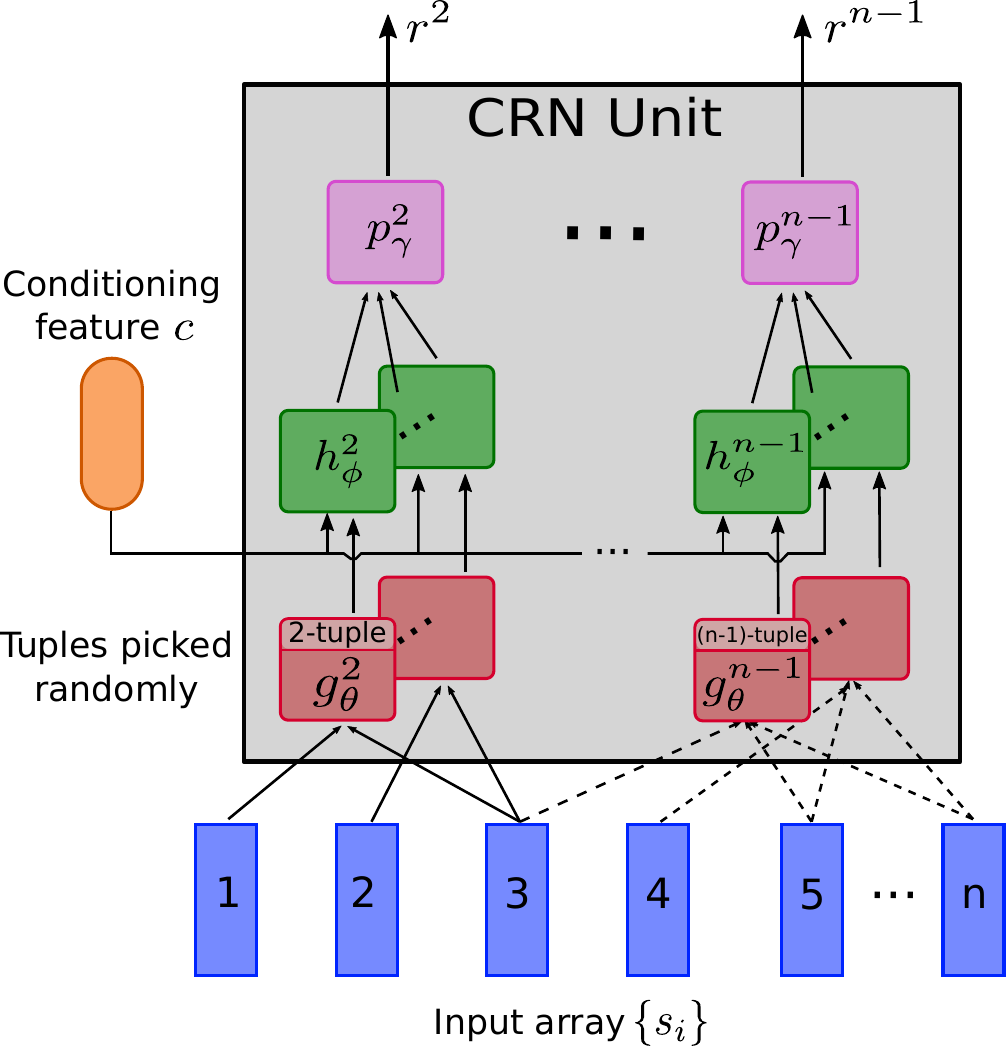}
\par\end{centering}
\caption{Conditional Relation Network. a) Input array {\small{}$\mathcal{S}$}
of $n$ objects are first processed to model $k$-tuple relations
from $t$ sub-sampled size-$k$ subsets by sub-network $g^{k}(.)$.
The outputs are further conditioned with the context $c$ via sub-network
$h^{k}(.,.)$ and finally aggregated by $p^{k}(.)$ to obtain a result
vector $r^{k}$ which represents $k$-tuple conditional relations.
Tuple sizes can range from $2$ to $(n-1)$, which outputs an $(n-2)$-dimensional
output array.\label{fig:Illustration-of-Multiscales}}
\medskip{}
\end{figure}

The goal of $\Problem$ is to deduce an answer $\tilde{a}$ from a
video $\mathcal{V}$ in response to a natural question $q$. The answer
$\tilde{a}$ can be found in an answer space $\mathcal{A}$ which
is a pre-defined set of possible answers for open-ended questions
or a list of answer candidates in case of multi-choice questions.
Formally, $\Problem$ can be formulated as follows:\vspace{-2mm}

\begin{equation}
\tilde{a}=\underset{a\in\mathcal{A}}{\text{argmax}}\mathcal{F}_{\theta}\left(a\mid q,\mathcal{V}\right),
\end{equation}
where $\theta$ is the model parameters of scoring function $\mathcal{F}$.\vspace{-3mm}

\paragraph{Visual representation}

We begin by dividing the video $\mathcal{V}$ of $L$ frames into
$N$ equal length clips $C=\left(C_{1},...,C_{N}\right)$. Each clip
$C_{i}$ of length $T=\left\lfloor L/N\right\rfloor $ is represented
by two sources of information: frame-wise appearance feature vectors
$V_{i}=\left\{ v_{i,j}|v_{i,j}\in\mathbb{R^{\text{2048}}}\right\} _{j=1}^{T}$
, and the motion feature vector at clip level $f_{i}\in\mathbb{R^{\text{2048}}}$.
In our experiments, $v_{i,j}$ are the \emph{pool5} output of ResNet
\cite{he2016deep} features and $f_{i}$ are derived by ResNeXt-101
\cite{xie2017aggregated,hara2018can}.

\begin{table}
\begin{centering}
\begin{tabular}{|l|l|}
\hline 
{\small{}Notation} & {\small{}Role}\tabularnewline
\hline 
\hline 
{\small{}$\mathcal{S}$} & {\small{}Input array of $n$ objects (e.g. frames, clips)}\tabularnewline
\hline 
{\small{}$c$} & {\small{}Conditioning feature (e.g. query, motion feat.)}\tabularnewline
\hline 
{\small{}$k_{\text{max}}$} & {\small{}Maximum subset (also tuple) size considered}\tabularnewline
\hline 
{\small{}$k$} & {\small{}Each subset size from $2$ to $k_{max}$}\tabularnewline
\hline 
{\small{}$Q^{k}$} & {\small{}Set of all size-$k$ subsets of $\mathcal{S}$}\tabularnewline
\hline 
{\small{}$t$} & {\small{}Number of subsets randomly selected from $Q^{k}$}\tabularnewline
\hline 
{\small{}$Q_{\textrm{selected}}^{k}$} & {\small{}Set of $t$ selected subsets from $Q^{k}$}\tabularnewline
\hline 
{\small{}$g^{k}(.)$} & {\small{}Sub-network processing each size-$k$ subset}\tabularnewline
\hline 
{\small{}$h^{k}(.,.)$} & {\small{}Conditioning sub-network}\tabularnewline
\hline 
{\small{}$p^{k}(.)$} & {\small{}Aggregating sub-network}\tabularnewline
\hline 
{\small{}$R$} & {\small{}Result array of CRN unit on $\mathcal{S}$ given $c$}\tabularnewline
\hline 
{\small{}$r^{k}$} & {\small{}Member result vector of $k$-tuple relations}\tabularnewline
\hline 
\end{tabular}\medskip{}
\par\end{centering}
\centering{}\caption{Notations of CRN unit operations\label{tab:notations}}
\end{table}

\begin{algorithm}[t]
\small
\label{algo:CRN}
\caption{CRN Unit}
	\SetKwInOut{Input}{Input}
	\SetKwInOut{Output}{Output}
	\SetKwInOut{Metaparams}{Metaparams}
	\Input{Array $\mathcal{S}=\{s_i\}_{i=1}^n$, conditioning feature $c$}
	\Output{Array $R$}
	\Metaparams{$\{k_{\textrm{max}},t \mid k_{\textrm{max}}<n\}$}
	Build all sets of subsets $\{Q^{k}\mid k=2,3,...,k_{\textrm{max}}\}$
where $Q^{k}$ is set of all size-$k$ subsets of $\mathcal{S}$

	Initialize $R \leftarrow \{\}$

	\For{$k \leftarrow 2$ \KwTo $k_{\SI{}{max}}$}
	{
		$Q^{k}_{\SI{}{selected}}=$ randomly select $t$ subsets from $Q^{k}$
		
		\For{\SI{}{\textbf{each}} subset $q_i$ $\in$ $Q^{k}_{\SI{}{selected}}$}
		{	
			$g_i = g^k(q_i)$
			
			$h_i = h^k(g_i,c)$
		}
        
		$r^k=p^{k}(\{h_i\})$
		
		add $r^k$ to $R$
		
	}
\end{algorithm}
\setlength{\textfloatsep}{6pt}

Subsequently, linear feature transformations are applied to project
$\{v_{ij}\}$ and $f_{i}$ into a standard $d$-dimensions feature
space to obtain $\{\hat{v}_{ij}|\hat{v}_{ij}\in\mathbb{R^{\text{d}}}\}$
and $\hat{f}_{i}\in\mathbb{R^{\text{d}}}$, respectively.\vspace{-5mm}

\paragraph{Linguistic representation}

All words in the question and answer candidates in case of multi-choice
questions are first embedded into vectors of 300 dimensions, which
are initialized with pre-trained GloVe word embeddings \cite{pennington2014glove}.
We further pass these context-independent embedding vectors through
a biLSTM. Output hidden states of the forward and backward LSTM passes
are finally concatenated to form the question representation $q\in\mathbb{R}^{d}$.

\begin{figure*}
\begin{centering}
\vspace{-4mm}
\includegraphics[width=0.78\textwidth,height=0.63\textwidth]{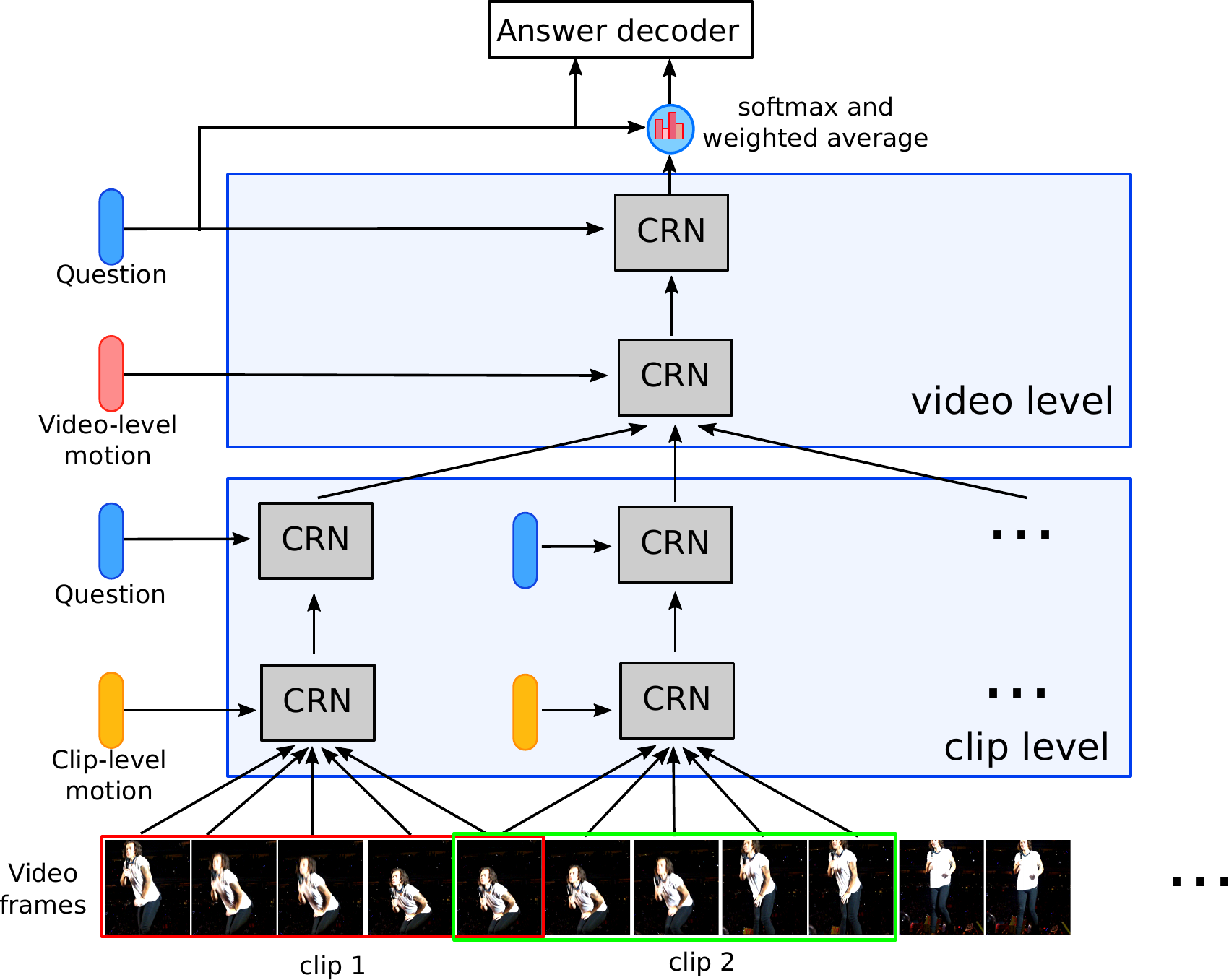}
\par\end{centering}
\caption{Hierarchical Conditional Relation Networks ($\protect\ModelName$)
Architecture for $\protect\Problem$. The $\protect\UnitName$s are
stacked in a hierarchy, embedding the video input at different granularities
including frame, short clip and entire video levels. The video feature
embedding is conditioned on the linguistic cue at each level of granularity.
The visual-question joint representation is fed into an output classifier
for prediction. \label{fig:Overview-of-Network}}
\medskip{}
\end{figure*}

With these representations, we now describe our new hierarchical architecture
for $\Problem$ (see Fig.~\ref{fig:Overview-of-Network}). We first
present the core compositional computation unit that serves as building
blocks for the architecture in Section~\ref{subsec:Multiscale-Relation-Network}.
In the following sub-section, we propose to design $\mathcal{F}$
as a layer-by-layer network architecture that can be built by simply
stacking the core units in a particular manner.

\subsection{Conditional Relation Network Unit\label{subsec:Multiscale-Relation-Network}}

We introduce a reusable computation unit, termed Conditional Relation
Network ($\UnitName$), which takes as input an array of $n$ objects
$\mathcal{S}=\{s_{i}\}_{i=1}^{n}$ and a conditioning feature $c$
- both in the same vector space $\mathbb{R}^{d}$ or tensor space
$\mathbb{R}^{W\times H\times d}$. CRN generates an output array of
objects of the same dimensions containing high-order object relations
of input features given the global context. The operation of CRN unit
is presented algorithmically in Alg.~\ref{algo:CRN} and visually
in Fig.~\ref{fig:Illustration-of-Multiscales}. Table \ref{tab:notations}
summarizes the notations used across these presentations.

When in use for $\Problem$, CRN's input array is composed of features
at either frame or short-clip levels. The objects $\{s_{i}\}_{i=1}^{n}$
greatly share mutual information and it is redundant to consider all
possible combinations of given objects. Therefore, applying a sampling
scheme on the set of subsets (line 4 of Alg.~\ref{algo:CRN}) is
crucial for redundancy reduction and computational efficiency. We
borrow the sampling trick in \cite{zhou2018temporal} to build sets
of $t$ selected subsets $Q_{\textrm{selected}}^{k}$. Regarding the
choice of $k_{\text{max}}$, we choose $k_{\text{max}}=n-1$ in later
experiments, resulting in the output array of size $n-2$ if $n>2$
and array of size $1$ if $n=2$.

As a choice in implementation, the functions $g^{k}(.),p^{k}(.)$
are simple average-pooling. In generic form, they can be any aggregation
sub-networks that join a random set into a single representation.
Meanwhile, $h^{k}(.,.)$ is a MLP running on top of feature concatenation
that models the non-linear relationships between multiple input modalities.
We tie parameters of the conditioning sub-network $h^{k}(.,.)$ across
the subsets of the same size $k$. In our implementation, $h^{k}(.,.)$
consists of a single linear transformation followed by an ELU \cite{clevert2015fast}
activation.

It may be of concern that the relation formed by a particular subset
may be unnecessary to model $k$-tuple relations, we optionally design
a self-gating mechanism similar to \cite{dauphin2017language} to
regulate the feature flow to go through each $\UnitName$ module.
Formally, the conditioning function $h^{k}(.,.)$ in that case is
given by:\vspace{-5mm}

\begin{equation}
h^{k}\left(x,y\right)=\text{ELU(}W_{h_{1}}\left[x,y\right])*\sigma\left(W_{h_{2}}\left[x,y\right]\right),\label{eq:gating_h}
\end{equation}
where $[.,.]$ denotes the tensor concatenation, $\sigma$ is sigmoid
function, and $W_{h_{1}},W_{h_{2}}$ are linear weights.

\subsection{Hierarchical Conditional Relation Networks \label{subsec:HCRN}}

We use $\UnitName$ blocks to build a deep network architecture to
exploit inherent characteristics of a video sequence namely temporal
relations, motion, and the hierarchy of video structure, and to support
reasoning guided by linguistic questions. We term the proposed network
architecture Hierarchical Conditional Relation Networks ($\ModelName$)
(see Fig.~\ref{fig:Overview-of-Network} ). The design of the $\ModelName$
by stacking reusable core units is partly inspired by modern CNN network
architectures, of which InceptionNet \cite{szegedy2015going} and
ResNet \cite{he2016deep} are the most well-known examples.

A model for $\Problem$ should distill the visual content in the context
of the question, given the fact that much of the visual information
is usually not relevant to the question. Drawing inspiration from
the hierarchy of video structure, we boil down the problem of $\Problem$
into a process of video representation in which a given video is encoded
progressively at different granularities, including short clip (consecutive
frames) and entire video levels. It is crucial that the whole process
conditions on linguistic cue. In particular, at each hierarchy level,
we use two stacked $\UnitName$ units, one conditioned on motion features
followed by one conditioned on linguistic cues. Intuitively, the motion
feature serves as a dynamic context shaping the temporal relations
found among frames (at the clip level) or clips (at the video level).
As the shaping effect is applied to all relations, self-gating is
not needed, and thus a simple MLP suffices. On the other hand, the
linguistic cues are by nature selective, that is, not all relations
are equally relevant to the question. Thus we utilize the self-gating
mechanism in Eq.~(\ref{eq:gating_h}) for the $\UnitName$ units
which condition on question representation. 

With this particular design of network architecture, the input array
at clip level consists of frame-wise appearance feature vectors $\{\hat{v}_{ij}\}$,
while that at a video level is the output at the clip level. Meanwhile,
the motion conditioning feature at clip level CRNs are corresponding
clip motion feature vector $\hat{f}_{i}$. They are further passed
to an LSTM, whose final state is used as video-level motion features.
Note that this particular implementation is not the only option. We
believe we are the first to progressively incorporate multiple modalities
of input in such a hierarchical manner in contrast to the typical
approach of treating appearance features and motion features as a
two-stream network.

To handle a long video of thousand frames, which is equivalent to
dozens of short-term clips, there are two options to reduce the computational
cost of $\UnitName$ in handling large sets of subsets $\{Q^{k}|k=2,3,...,k_{\text{max}}\}$
given an input array $S$: limit the maximum subset size $k_{\text{max}}$
or extend the $\ModelName$ to deeper hierarchy. For the former option,
this choice of sparse sampling may have potential to lose critical
relation information of specific subsets. The latter, on the other
hand, is able to densely sample subsets for relation modeling. Specifically,
we can group $N$ short-term clips into $N_{1}\times N_{2}$ hyper-clips,
of which $N_{1}$ is the number of the hyper-clips and $N_{2}$ is
the number of short-term clips in one hyper-clip. By doing this, our
$\ModelName$ now becomes a 3-level of hierarchical network architecture.

At the end of the $\ModelName$, we compute the average visual feature
based on conditioning to the question representation $q$. Assume
outputs of the last $\UnitName$ unit at video level are an array
$O=\left\{ o_{i}\mid o_{i}\in\mathbb{R}^{H\times d}\right\} _{i=1}^{N-4}$,
we first stack them together, resulting in an output tensor $o\in\mathbb{R}^{(N-4)\times H\times d}$,
and further vectorize this output tensor to obtain the final output
$o^{\prime}\in\mathbb{R}^{H^{\prime}\times d},H^{\prime}=(N-4)\times H$.
The weighted average information is given by:

\vspace{-3mm}

\begin{align}
I & =\left[W_{o^{\prime}}o^{\prime},W_{o^{\prime}}o^{\prime}\odot W_{q}q\right],\\
I^{\prime} & =\text{ELU}\left(W_{I}I+b\right),\\
\gamma & =\text{softmax}\left(W_{I^{\prime}}I^{\prime}+b\right),\\
\tilde{o} & =\sum_{h=1}^{H^{\prime}}\gamma_{h}o_{h}^{\prime};\,\tilde{o}\in\mathbb{R}^{d},
\end{align}
where, $\left[.,.\right]$ denotes concatenation operation, and $\odot$
is the Hadamard product.

\subsection{Answer Decoders and Loss Functions\label{subsec:Answer-Decoders}}

Following \cite{jang2017tgif,song2018explore,fan2019heterogeneous},
we adopt different answer decoders depending on the task. Open-ended
questions are treated as multi-label classification problems. For
these, we employ a classifier which takes as input the combination
of the retrieved information from visual cue $\tilde{o}$ and the
question representation $q$, and computes label probabilities $p\in\mathbb{R}^{|\mathcal{A}|}$:\vspace{-1.3mm}

\begin{align}
y & =\text{ELU}\left(W_{o}\left[\tilde{o},W_{q}q+b\right]+b\right),\\
y^{\prime} & =\text{ELU}\left(W_{y}y+b\right),\label{eq:9}\\
p & =\textrm{softmax}\left(W_{y^{\prime}}y^{\prime}+b\right).
\end{align}
The cross-entropy is used as the loss function.

For repetition count task, we use a linear regression function taking
$y^{\prime}$ in Eq.~(\ref{eq:9}) as input, followed by a rounding
function for integer count results. The loss for this task is Mean
Squared Error (MSE).

For multi-choice question types (such as repeating action and state
transition in TGIF-QA), each answer candidate is processed in the
same way with the question. In detail, we use the shared parameter
$\ModelName$s with either question or each answer candidate as language
cues. As a result, we have a set of $\ModelName$ outputs, one conditioned
on question ($\tilde{o}_{q}$), and the others conditioned on answer
candidates ($\tilde{o}_{a})$. Subsequently, $\tilde{o}_{q},$ $\{\tilde{o}_{a}\}$,
question representation $q$ and answer candidates $a$ are fed into
a final classifier with a linear regression to output an answer index,
as follows:\vspace{-1.3mm}
\begin{align}
y & =\left[\tilde{o}_{q},\tilde{o}_{a},W_{q}q+b,W_{a}a+b\right],\\
y^{\prime} & =\text{ELU}\left(W_{y}y+b\right),\\
s & =W_{y^{\prime}}y^{\prime}+b.
\end{align}
We use the popular hinge loss \cite{jang2017tgif} of pairwise comparisons,
$\text{max}\left(0,1+s^{n}-s^{p}\right)$, between scores for incorrect
$s^{n}$ and correct answers $s^{p}$ to train the network.

\subsection{Complexity Analysis \label{subsec:Complexity-Analysis}}

We provide a brief analysis here, leaving detailed derivations in
Supplement. For a fixed sampling resolution $t$, a single forward
pass of $\UnitName$ would take quadratic time in $k_{\text{max}}$.
For an input array of length $n$, feature size $F$, the unit produces
an output array of size $k_{\text{max}}-1$ of the same feature dimensions.
The overall complexity of $\ModelName$ depends on design choice for
each $\UnitName$ unit and specific arrangement of $\UnitName$ units.
For clarity, let $t=2$ and $k_{\max}=n-1$, which are found to work
well in later experiments. Suppose there are $N$ clips of length
$T$, making a video of length $L=NT$. A 2-level architecture of
Fig.~\ref{fig:Overview-of-Network} needs $2TLF$ time to compute
the $\UnitName$s at the lowest level, and $2NLF$ time to compute
the second level, totaling $2(T+N)LF$ time.

Let us now analyze a 3-level architecture that generalizes the one
in Fig.~\ref{fig:Overview-of-Network}. The $N$ clips are organized
into $M$ sub-videos, each has $Q$ clips, i.e., $N=MQ$. The clip-level
$\UnitName$s remain the same. At the next level, each sub-video $\UnitName$
takes as input an array of length $Q$, whose elements have size $(T-4)F$.
Using the same logic as before, the set of sub-video-level CRNs cost
$2\frac{N}{M}LF$ time. A stack of two sub-video $\UnitName$s now
produces an output array of size $(Q-4)(T-4)F$, serving as an input
object in an array of length $M$ for the video-level $\UnitName$s.
Thus the video-level $\UnitName$s cost $2MLF$. Thus the total cost
for 3-level $\ModelName$ is in the order of $2(T+\frac{N}{M}+M)LF$.

Compared to the 2-level $\ModelName$, the a 3-level $\ModelName$
reduces computation time by $2(N-\frac{N}{M}-M)LF\approx2NLF$ assuming
$N\gg\max\left\{ M,\frac{N}{M}\right\} $. As $N=\frac{L}{T}$, this
reduces to $2NLF=2\frac{L^{2}}{T}F$. In practice $T$ is often fixed,
thus the saving scales quadratically with video length $L$, suggesting
that hierarchy is computational efficient for long videos.

\section{Experiments \label{sec:Experiments}}

\subsection{Datasets}

\paragraph*{TGIF-QA \cite{jang2017tgif}}

This is currently the most prominent dataset for $\Problem$, containing
165K QA pairs and 72K animated GIFs. The dataset covers four tasks
addressing unique properties of video. Of which, the first three
require strong spatio-temporal reasoning abilities\emph{: Repetition
Count} - to retrieve number of occurrences of an action, \emph{Repeating
Action}- multi-choice task to identify the action repeated for a
given number of times, \emph{State Transition} - multi-choice tasks
regarding temporal order of events. The last task - \emph{Frame QA}
- is akin to image QA where a particular frame in a video is sufficient
to answer the questions.\vspace{-4mm}

\paragraph{MSVD-QA \cite{xu2017video}}

This is a small dataset of 50,505 question answer pairs annotated
from 1,970 short clips. Questions are of five types, including what,
who, how, when and where.

\vspace{-4mm}

\paragraph*{MSRVTT-QA \cite{xu2016msr}}

The dataset contains 10K videos and 243K question answer pairs. Similar
to MSVD-QA, questions are of five types. Compared to the other two
datasets, videos in MSRVTT-QA contain more complex scenes. They are
also much longer, ranging from 10 to 30 seconds long, equivalent to
300 to 900 frames per video.

We use accuracy to be the evaluation metric for all experiments, except
those for repetition count on TGIF-QA dataset where Mean Square Error
(MSE) is applied.

\subsection{Implementation Details}

Videos are segmented into 8 clips, each clip contains 16 frames by
default. Long videos in MSRVTT-QA are additionally segmented into
24 clips for evaluating the ability of handling very long sequences.
Unless otherwise stated, the default setting is with a 2-level $\ModelName$
as depicted in Fig.~\ref{fig:Overview-of-Network}, and $d=512$,
$t=1$. We train the model initially at learning rate of $10^{-4}$
and decay by half after every 10 epochs. All experiments are terminated
after 25 epochs and reported results are at the epoch giving the best
validation accuracy. Pytorch implementation of the model is available
online\footnote{https://github.com/thaolmk54/hcrn-videoqa}.

\begin{table}
\begin{centering}
\begin{tabular}{l|c|c|c|c}
\hline 
Model & Action & Trans. & Frame & Count\tabularnewline
\hline 
ST-TP \cite{jang2017tgif} & 62.9 & 69.4 & 49.5 & 4.32\tabularnewline
Co-mem \cite{gao2018motion} & 68.2 & 74.3 & 51.5 & 4.10\tabularnewline
PSAC \cite{li2019beyond} & 70.4 & 76.9 & 55.7 & 4.27\tabularnewline
HME \cite{fan2019heterogeneous} & 73.9 & 77.8 & 53.8 & 4.02\tabularnewline
\textbf{HCRN} & \textbf{75.0} & \textbf{81.4} & \textbf{55.9} & \textbf{3.82}\tabularnewline
\hline 
\end{tabular}\medskip{}
\par\end{centering}
\caption{Comparison with the state-of-the-art methods on TGIF-QA dataset. For
count, the lower the better.\label{tab:tgif}}
\end{table}

\begin{figure}
\begin{centering}
\includegraphics[width=0.8\columnwidth,height=0.63\columnwidth]{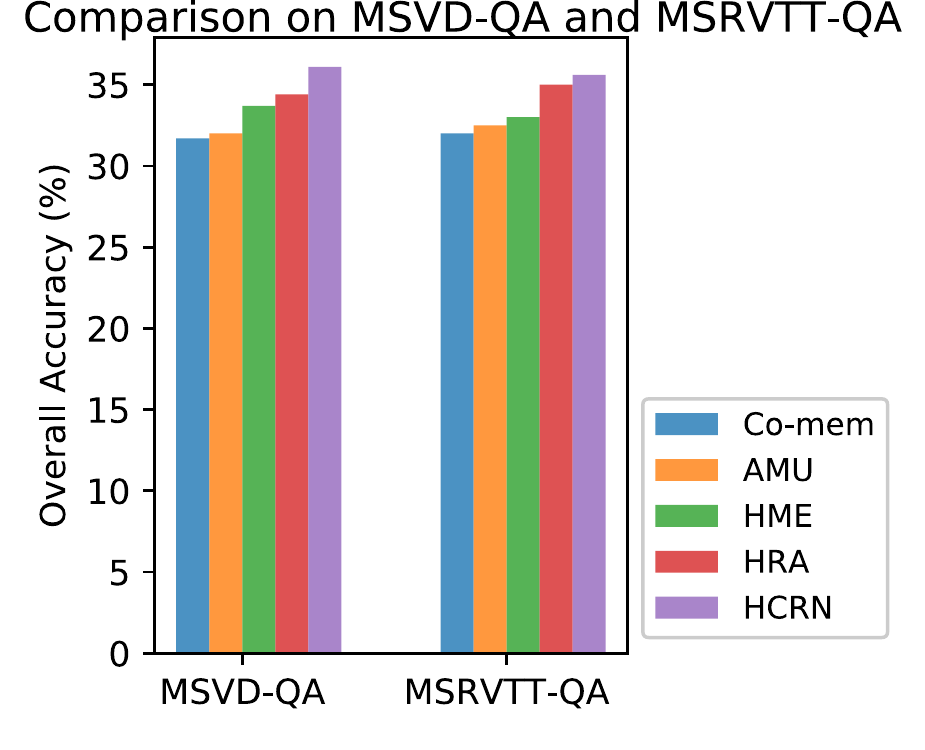}\vspace{-2mm}
\par\end{centering}
\caption{Performance comparison on MSVD-QA and MSRVTT-QA dataset with state-of-the-art
methods: Co-mem \cite{gao2018motion}, HME \cite{fan2019heterogeneous},
HRA \cite{chowdhury2018hierarchical}, and AMU \cite{xu2017video}.
\label{fig:msvd-msrvtt}}
\medskip{}
\end{figure}

\subsection{Results}

\subsubsection{Benchmarking against SoTAs}

We compare our proposed model with state-of-the-art methods (SoTAs)
on aforementioned datasets. For TGIF-QA, we compare with most recent
SoTAs, including \cite{fan2019heterogeneous,gao2018motion,jang2017tgif,li2019beyond},
over four tasks. These works, except for \cite{li2019beyond}, make
use of motion features extracted from optical flow or 3D CNNs.

The results are summarized in Table~\ref{tab:tgif} for TGIF-QA,
and in Fig.~\ref{fig:msvd-msrvtt} for MSVD-QA and MSRVTT-QA. Reported
numbers of the competitors are taken from the original papers and
\cite{fan2019heterogeneous}. It is clear that our model consistently
outperforms or is competitive with SoTA models on all tasks across
all datasets. The improvements are particularly noticeable when strong
temporal reasoning is required, i.e., for the questions involving
actions and transitions in TGIF-QA. These results confirm the significance
of considering both near-term and far-term temporal relations toward
finding correct answers. 

The MSVD-QA and MSRVTT-QA datasets represent highly challenging benchmarks
for machine compared to the TGIF-QA, thanks to their open-ended nature.
Our model $\ModelName$ outperforms existing methods on both datasets,
achieving 36.1\% and 35.6\% accuracy which are 1.7 points and 0.6
points improvement on MSVD-QA and MSRVTT-QA, respectively. This suggests
that the model can handle both small and large datasets better than
existing methods. 

Finally, we provide a justification for the competitive performance
of our $\ModelName$ against existing rivals by comparing model features
in Table~\ref{tab:Model-design-choices}. Whilst it is not straightforward
to compare head-to-head on internal model designs, it is evident that
effective video modeling necessitates handling of motion, temporal
relation and hierarchy at the same time. We will back this hypothesis
by further detailed studies in Section~\ref{subsec:Ablation-Studies}
(for motion, temporal relations, shallow hierarchy) and Section~\ref{subsec:Deepening-model-hierarchy}
(deep hierarchy).

\begin{table}
\begin{centering}
\vspace{-1mm}
{\scriptsize{}}%
\begin{tabular}{l|c|c|c|c}
\hline 
{\small{}Model} & {\small{}Appear.} & {\small{}Motion} & {\small{}Hiera.} & {\small{}Relation}\tabularnewline
\hline 
{\small{}ST-TP \cite{jang2017tgif}} & {\small{}$\checkmark$} & {\small{}$\checkmark$} &  & \tabularnewline
{\small{}Co-mem \cite{gao2018motion}} & {\small{}$\checkmark$} & {\small{}$\checkmark$} &  & \tabularnewline
{\small{}PSAC \cite{li2019beyond}} & {\small{}$\checkmark$} &  &  & \tabularnewline
{\small{}HME \cite{fan2019heterogeneous}} & {\small{}$\checkmark$} & {\small{}$\checkmark$} &  & \tabularnewline
{\small{}$\ModelName$} & {\small{}$\checkmark$} & {\small{}$\checkmark$} & {\small{}$\checkmark$} & {\small{}$\checkmark$}\tabularnewline
\hline 
\end{tabular}\medskip{}
\par\end{centering}
\caption{Model design choices and input modalities in comparison. See Table~\ref{tab:tgif}
for corresponding performance on TGIF-QA dataset. \label{tab:Model-design-choices}}
\medskip{}
\end{table}

\subsubsection{Ablation Studies \label{subsec:Ablation-Studies}}

\begin{table}
\centering{}%
\begin{tabular}{l|c|c|c|c}
\hline 
{\small{}Model} & {\small{}Act.} & {\small{}Trans.} & {\small{}F.QA} & {\small{}Count}\tabularnewline
\hline 
\hline 
\textbf{\small{}Relations $(k_{max},t)$} &  &  &  & \tabularnewline
{\small{}$\quad$$k_{max}=1,t=1$} & {\small{}65.2} & {\small{}75.5} & {\small{}54.9} & {\small{}3.97}\tabularnewline
{\small{}$\quad$$k_{max}=1,t=3$} & {\small{}66.2} & {\small{}76.2} & {\small{}55.7} & {\small{}3.95}\tabularnewline
{\small{}$\quad$$k_{max}=1,t=5$} & {\small{}65.4} & {\small{}76.7} & {\small{}56.0} & {\small{}3.91}\tabularnewline
{\small{}$\quad$$k_{max}=1,t=9$} & {\small{}65.6} & {\small{}75.6} & {\small{}56.3} & {\small{}3.92}\tabularnewline
{\small{}$\quad$$k_{max}=1,t=11$} & {\small{}65.4} & {\small{}75.1} & {\small{}56.3} & {\small{}3.91}\tabularnewline
{\small{}$\quad$$k_{max}=2,t=2$} & {\small{}67.2} & {\small{}76.6} & {\small{}56.7} & {\small{}3.94}\tabularnewline
{\small{}$\quad$$k_{max}=2,t=9$} & {\small{}66.3} & {\small{}76.7} & {\small{}56.5} & {\small{}3.92}\tabularnewline
{\small{}$\quad$$k_{max}=4,t=2$} & {\small{}64.0} & {\small{}75.9} & {\small{}56.2} & {\small{}3.87}\tabularnewline
{\small{}$\quad$$k_{max}=4,t=9$} & {\small{}66.3} & {\small{}75.6} & {\small{}55.8} & {\small{}4.00}\tabularnewline
{\small{}$\quad$$k_{max}=\left\lfloor n/2\right\rfloor ,t=2$} & {\small{}73.3} & {\small{}81.7} & {\small{}56.1} & {\small{}3.89}\tabularnewline
{\small{}$\quad$$k_{max}=\left\lfloor n/2\right\rfloor ,t=9$} & {\small{}72.5} & {\small{}81.1} & {\small{}56.6} & {\small{}3.82}\tabularnewline
{\small{}$\quad$$k_{max}=n-1,t=1$} & {\small{}75.0} & {\small{}81.4} & {\small{}55.9} & {\small{}3.82}\tabularnewline
{\small{}$\quad$$k_{max}=n-1,t=3$} & {\small{}75.1} & {\small{}81.5} & {\small{}55.5} & {\small{}3.91}\tabularnewline
{\small{}$\quad$$k_{max}=n-1,t=5$} & {\small{}73.6} & {\small{}82.0} & {\small{}54.7} & {\small{}3.84}\tabularnewline
{\small{}$\quad$$k_{max}=n-1,t=7$} & {\small{}75.4} & {\small{}81.4} & {\small{}55.6} & {\small{}3.86}\tabularnewline
{\small{}$\quad$$k_{max}=n-1,t=9$} & {\small{}74.1} & {\small{}81.9} & {\small{}54.7} & {\small{}3.87}\tabularnewline
\hline 
\textbf{\small{}Hierarchy} &  &  &  & \tabularnewline
{\small{}$\quad$$1$-level, video $\UnitName$ only} & {\small{}66.2} & {\small{}78.4} & {\small{}56.6} & {\small{}3.94}\tabularnewline
{\small{}$\quad$$1.5$-level, clips$\rightarrow$pool} & {\small{}70.4} & {\small{}80.5} & {\small{}56.6} & {\small{}3.94}\tabularnewline
\hline 
\textbf{\small{}Motion conditioning} &  &  &  & \tabularnewline
{\small{}$\quad$w/o motion} & {\small{}70.8} & {\small{}79.8} & {\small{}56.4} & {\small{}4.38}\tabularnewline
{\small{}$\quad$w/o short-term motion} & {\small{}74.9} & {\small{}82.1} & {\small{}56.5} & {\small{}4.03}\tabularnewline
{\small{}$\quad$w/o long-term motion} & {\small{}75.1} & {\small{}81.3} & {\small{}56.7} & {\small{}3.92}\tabularnewline
\hline 
\textbf{\small{}Linguistic conditioning} &  &  &  & \tabularnewline
{\small{}$\quad$w/o linguistic condition} & {\small{}66.5} & {\small{}75.7} & {\small{}56.2} & {\small{}3.97}\tabularnewline
{\small{}$\quad$w/o quest.@clip level} & {\small{}74.3} & {\small{}81.1} & {\small{}55.8} & {\small{}3.95}\tabularnewline
{\small{}$\quad$w/o quest.@video level} & {\small{}74.0} & {\small{}80.5} & {\small{}55.9} & {\small{}3.92}\tabularnewline
\hline 
\textbf{\small{}Gating} &  &  &  & \tabularnewline
{\small{}$\quad$w/o gate} & {\small{}74.1} & {\small{}82.0} & {\small{}55.8} & {\small{}3.93}\tabularnewline
{\small{}$\quad$w/ gate quest. \& motion} & {\small{}73.3} & {\small{}80.9} & {\small{}55.3} & {\small{}3.90}\tabularnewline
\hline 
{\small{}$\quad$Full $2$-level $\ModelName$} & {\small{}75.1} & {\small{}81.2} & {\small{}55.7} & {\small{}3.88}\tabularnewline
\hline 
\end{tabular}\medskip{}
\caption{Ablation studies on TGIF-QA dataset. For count, the lower the better.
Act.: Action; Trans.: Transition; F.QA: Frame QA. When not explicitly
specified, we use $k_{max}=n-1,t=2$ for relation order and sampling
resolution. \label{tab:Ablation-tgif}}
\medskip{}
\end{table}

To provide more insight about our model, we conduct extensive ablation
studies on TGIF-QA with a wide range of configurations.\emph{ }The
results are reported in Table~\ref{tab:Ablation-tgif}. \emph{Full
$2$-level $\ModelName$} denotes the full model of Fig.~\ref{fig:Overview-of-Network}
with $k_{max}=n-1,t=2$. Overall we find that ablating any of design
components or $\UnitName$ units would degrade the performance for
temporal reasoning tasks (actions, transition and action counting).
The effects are detailed as follows.\vspace{-3mm}

\paragraph{Effect of relation order $k_{max}$ and resolution $t$}

Without relations ($k_{max}=1$) the performance drops significantly
on actions and events reasoning. This is expected since those questions
often require putting actions and events in relation with a larger
context (e.g., what happens before something else). In this case,
the frame QA benefits more from increasing sampling resolution $t$
because of better chance to find a relevant frame. However, when taking
relations into account ($k_{max}>1$), we find that $\ModelName$
is robust against sampling resolution $t$ but depends critically
on the maximium relation order $k_{max}$. The relative independence
w.r.t. $t$ can be due to visual redundancy between frames, so that
resampling may capture almost the same information. On the other hand,
when considering only low-order object relations, the performance
is significantly dropped in all tasks, except frame QA. These results
confirm that high-order relations are required for temporal reasoning.
As the frame QA task requires only reasoning on a single frame, incorporating
temporal information might confuse the model.\vspace{-5mm}

\paragraph{Effect of hierarchy}

We design two simpler models with only one $\UnitName$ layer: \emph{$\blacktriangleright$
}$1$\emph{-level, $1$ $\UnitName$ video on key frames only}: Using
only one $\UnitName$ at the video-level whose input array consists
of key frames of the clips. Note that video-level motion features
are still maintained. \emph{$\blacktriangleright$ }$1.5$-\emph{level,
clip $\UnitName$s $\rightarrow$} \emph{pooling}: Only the clip-level
$\UnitName$s are used, and their outputs are mean-pooled to represent
video. The pooling operation represents a simplistic relational operation
across clips. The results confirm that a hierarchy is needed for high
performance on temporal reasoning tasks.\vspace{-3mm}

\paragraph{Effect of motion conditioning}

We evaluate the following settings: \emph{$\blacktriangleright$ w/o
short-term motions}: Remove all $\UnitName$ units that condition
on the short-term motion features (clip level) in the $\ModelName$.
\emph{$\blacktriangleright$ w/o long-term motions}: Remove the $\UnitName$
unit that conditions on the long-term motion features (video level)
in the $\ModelName$.\emph{ $\blacktriangleright$ w/o motions}: Remove
motion feature from being used by $\ModelName$. We find that motion,
in agreeing with prior arts, is critical to detect actions, hence
computing action count. Long-term motion is particularly significant
for counting task, as this task requires maintaining global temporal
context during the entire process. For other tasks, short-term motion
is usually sufficient. E.g. in action task, wherein one action is
repeatedly performed during the entire video, long-term context contributes
little. Not surprisingly, motion does not play the positive role in
answering questions on single frames as only appearance information
needed.\vspace{-3mm}

\paragraph{Effect of linguistic conditioning and gating}

Linguistic cues represent a crucial context for selecting relevant
visual artifacts. For that we test the following ablations:\emph{
$\blacktriangleright$ w/o quest.@clip level}: Remove the $\UnitName$
unit that conditions on question representation at clip level. \emph{$\blacktriangleright$
w/o quest.@video level}: Remove the $\UnitName$ unit that conditions
on question representation at video level. \emph{$\blacktriangleright$
w/o linguistic condition: }Exclude all $\UnitName$ units conditioning
on linguistic cue while the linguistic cue is still in the answer
decoder. Likewise, gating offers a selection mechanism. Thus we study
its effect as follows: \emph{$\blacktriangleright$ wo/ gate}: Turn
off the self-gating mechanism in all $\UnitName$ units.\emph{ $\blacktriangleright$
w/ gate quest. \& motion}: Turn on the self-gating mechanism in all
$\UnitName$ units. 

We find that the conditioning question provides an important context
for encoding video. Conditioning features (motion and language), through
the gating mechanism in Eq.~(\ref{eq:gating_h}), offers further
performance gain in action and counting tasks, possibly by selectively
passing question-relevant information up the inference chain.

\begin{table}
\begin{centering}
\begin{tabular}{l|c}
\hline 
Depth of hierarchy & Overall Acc.\tabularnewline
\hline 
$2$-level, $24$ clips $\rightarrow$ $1$ vid & 35.6\tabularnewline
$3$-level, $24$ clips $\rightarrow$ $4$ sub-vids $\rightarrow$
$1$ vid & 35.6\tabularnewline
\hline 
\end{tabular}\medskip{}
\par\end{centering}
\caption{Results for going deeper hierarchy on MSRVTT-QA dataset. Run time
is reduced by factor of $4$ for going from 2-level to 3-level hierarchy.
\label{tab:scale_analysis-msrvtt}}
\medskip{}
\end{table}

\subsubsection{Deepening model hierarchy \label{subsec:Deepening-model-hierarchy}}

We test the scalability of the HCRN on long videos in the MSRVTT-QA
dataset, which are organized into 24 clips (3 times longer than other
two datasets). We consider two settings: \emph{$\blacktriangleright$
$2$-level hierarchy,} \emph{$24$ clips}$\rightarrow$$1$ \emph{vid}:
The model is as illustrated in Fig.~\ref{fig:Overview-of-Network},
where 24 clip-level $\UnitName$s are followed by a video-level $\UnitName$.
\emph{$\blacktriangleright$ $3$-level hierarchy}, $24$ \emph{clips}$\rightarrow$$4$\emph{
sub-vid}s$\rightarrow$$1$ \emph{vid}: Starting from the 24 clips
as in the $2$-level hierarchy, we group 24 clips into 4 sub-videos,
each is a group of 6 consecutive clips, resulting in a $3$-level
hierarchy. These two models are designed to have similar number of
parameters, approx. 50M.

The results are reported in Table~\ref{tab:scale_analysis-msrvtt}.
Unlike existing methods which usually struggle with handling long
videos, our method is scalable for them by offering deeper hierarchy,
as analyzed theoretically in Section~\ref{subsec:Complexity-Analysis}.
Using a deeper hierarchy is expected to significantly reduce the training
time and inference time for $\ModelName$, especially when the video
is long. In our experiments, we achieve \emph{4 times reduction in
training and inference time} by going from 2-level $\ModelName$ to
3-level counterpart whilst maintaining the same performance.

\section{Discussion \label{sec:Discussion}}

We introduced a general-purpose neural unit called Conditional Relational
Networks (CRNs) and a method to construct hierarchical networks for
$\Problem$ using $\UnitName$s as building blocks. A CRN is a relational
transformer that encapsulates and maps an array of tensorial objects
into a new array of the same kind, conditioned on a contextual feature.
In the process, high-order relations among input objects are encoded
and modulated by the conditioning feature. This design allows flexible
construction of sophisticated structure such as stack and hierarchy,
and supports iterative reasoning, making it suitable for QA over multimodal
and structured domains like video. The $\ModelName$ was evaluated
on multiple $\Problem$ datasets (TGIF-QA, MSVD-QA, MSRVTT-QA) demonstrating
competitive reasoning capability.

Different to temporal attention based approaches which put effort
into selecting objects, HCRN concentrates on modeling relations and
hierarchy in video. This difference in methodology and design choices
leads to distinctive benefits. CRN units can be further augmented
with attention mechanisms to cover better object selection ability,
so that related tasks such as frame QA can be further improved.

The examination of $\UnitName$ in $\Problem$ highlights the importance
of building generic neural reasoning unit that supports native multimodal
interaction in improving robustness of visual reasoning. We wish to
emphasize that the unit is general-purpose, and hence is applicable
for other reasoning tasks, which we will explore. These includes an
extension to consider the accompanying linguistic channels which are
crucial for TVQA \cite{lei2018tvqa} and MovieQA \cite{tapaswi2016movieqa}
tasks.

{\small{}\bibliographystyle{ieee}
\bibliography{../../bibs/thaole,../../bibs/truyen,../../bibs/ME,../../bibs/video}
}{\small\par}
\end{document}